\documentclass[conference]{IEEEtran}
\IEEEoverridecommandlockouts
\usepackage{caption}

\usepackage{cite}
\usepackage{amsmath,amssymb,amsfonts,amsthm}
\usepackage{algorithmic}
\usepackage{graphicx}
\usepackage{svg}
\usepackage{textcomp}
\usepackage{comment}
\usepackage{booktabs} 
\usepackage{multirow} 
\usepackage{xurl}
\usepackage{bm}
\usepackage{acro}
\usepackage{tikz}
\usetikzlibrary{shapes.geometric, arrows, positioning, fit}

\DeclareAcronym{ODD}{
    short = ODD, 
    long = operational design domain
}

\DeclareAcronym{ADS}{
    short = ADS, 
    long = automated driving system
}

\DeclareAcronym{WTTC}{
    short = WTTC, 
    long = Worst-Time-To-Collision
}

\DeclareAcronym{TTC}{
    short = TTC, 
    long = Time-To-Collision
}
 
\DeclareAcronym{HiL}{
    short = HiL, 
    long = hardware-in-the-loop
}

\DeclareAcronym{SiL}{
    short = SiL, 
    long = software-in-the-loop
}

\DeclareAcronym{MiL}{
    short = MiL, 
    long = model-in-the-loop
}

\DeclareAcronym{XiL}{
    short = XiL, 
    long = x-in-the-loop
}

\DeclareAcronym{SuT}{
    short = SuT, 
    long = system under test
}

\DeclareAcronym{PG}{
    short = PG, 
    long = Proving Ground
}

\DeclareAcronym{FOT}{
    short = FOT, 
    long = Field Operational Test
}

\DeclareAcronym{ECU}{
    short = ECU, 
    long = electronic control unit
}

\DeclareAcronym{ACC}{
    short = ACC, 
    long = adaptive cruise control
}

\DeclareAcronym{AEBS}{
    short = AEBS, 
    long = advanced emergency braking system
}

\DeclareAcronym{HAD}{
    short = HAD, 
    long = highly automated driving
}

\DeclareAcronym{ADAS}{
    short = ADAS, 
    long = advanced driver-assistance systems
}

\DeclareAcronym{ML}{
    short = ML, 
    long = machine learning
}

\DeclareAcronym{LKA}{
    short = LKA,
    long = lane keeping assist
}

\DeclareAcronym{ASIL}{
    short = ASIL,
    long = automotive safety integrity level
}

\DeclareAcronym{AV}{
    short = AV,
    long = automated vehicle
}

\DeclareAcronym{SAE}{
    short = SAE,
    long = Society of Automotive Engineers
}

\DeclareAcronym{IMU}{
    short = IMU,
    long = inertial measurement unit
}

\DeclareAcronym{ssim}{
    short = SSIM,
    long = Structural Similarity
}

\DeclareAcronym{lpips}{
    short = LPIPS,
    long = Learned Perceptual Image Patch Similarity
}

\DeclareAcronym{psnr}{
    short = PSNR,
    long = Peak Signal to Noise Ratio
}

\DeclareAcronym{mse}{
    short = MSE,
    long = Mean Squared Error
}

\DeclareAcronym{fid}{
    short = FID,
    long = Fr\'echet Inception Distance
}

\DeclareAcronym{mi}{
    short = MI,
    long = Mutual Information
}

\DeclareAcronym{ncc}{
    short = NCC,
    long = Normalized Cross-Correlation
}

\DeclareAcronym{vqgan}{
    short = VQGAN,
    long = Vector Quantized Generative Adversarial Networks
}

\usepackage[hidelinks]{hyperref}
\def\BibTeX{{\rm B\kern-.05em{\sc i\kern-.025em b}\kern-.08em
    T\kern-.1667em\lower.7ex\hbox{E}\kern-.125emX}}
\begin{document}
\UseRawInputEncoding
\title{Data Quality Matters: Quantifying Image Quality Impact on Machine Learning Performance
}
\author{\IEEEauthorblockN{Christian Steinhauser$^{*}$, Philipp Reis$^{*}$, Hubert Padusinski, Jacob Langner and Eric Sax}
\IEEEauthorblockA{
FZI Research Center for Information Technology, Karlsruhe, Germany\\
Email: \{steinhauser, reis, padusinski, langner, sax\}@fzi.de}
\thanks{$^{*}$ Both authors contributed equally to this work.}
}
\theoremstyle{definition}
\newtheorem{definition}{Definition}[section]
\maketitle
\bibliographystyle{IEEEtran}
\begin{abstract}
Precise perception of the environment is essential in highly automated driving systems, which rely on machine learning tasks such as object detection and segmentation. Compression of sensor data is commonly used for data handling, while virtualization is used for hardware-in-the-loop validation. Both methods can alter sensor data and degrade model performance. This necessitates a systematic approach to quantifying image validity.
This paper presents a four-step framework to evaluate the impact of image modifications on machine learning tasks. First, a dataset with modified images is prepared to ensure one-to-one matching image pairs, enabling measurement of deviations resulting from compression and virtualization. Second, image deviations are quantified by comparing the effects of compression and virtualization against original camera-based sensor data. Third, the performance of state-of-the-art object detection models is analyzed to determine how altered input data affects perception tasks, including bounding box accuracy and reliability. Finally, a correlation analysis is performed to identify relationships between image quality and model performance. As a result, the LPIPS metric achieves the highest correlation between image deviation and machine learning performance across all evaluated machine learning tasks.  
\end{abstract}

\section{Introduction}
\begin{figure}[t!]
    \centering
    \includegraphics[width=\linewidth]{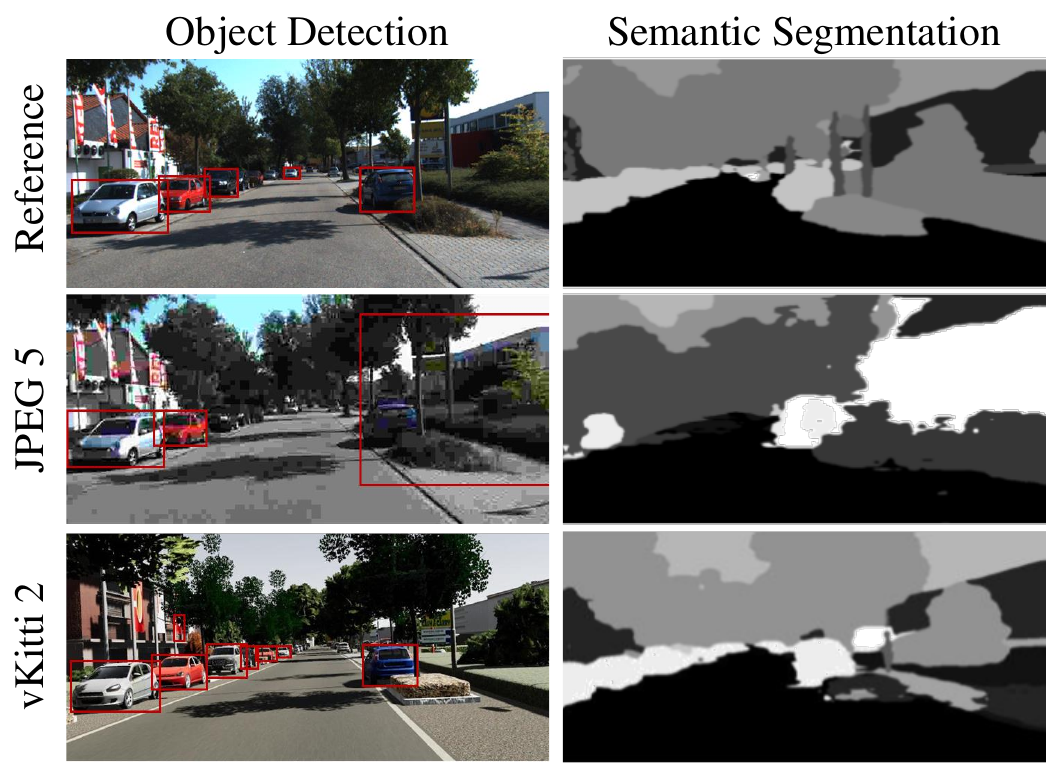}
    \caption{Compression (JPEG~5) or virtualization (vKitti~2) can lead to deviating results in machine learning tasks. The object detection (left) shows false positives compared to the reference. Semantic segmentation (right) shows misclassifications. Our approach aims to quantify deviations in image quality in relation to ML-performance.}
    \label{fig:Cachy_Figure}
\end{figure}

The advent of highly automated driving systems has introduced numerous challenges, with environmental perception being crucial for ensuring safe and efficient vehicle operation. The increasing reliance on \ac{ML} methods for tasks such as object detection and segmentation in vehicles highlights the need for high-quality input data to maintain optimal performance. 
The validation and verification of automated driving functions is carried out on the basis of two distinct data sources: Recorded real data and simulated data~\cite{rigoll_focus_2023, steinhauser_efficient_2022}. In both cases, deviations may be observed when compared to the raw data recorded in the vehicle (see Figure~\ref{fig:Cachy_Figure}). In the simulation, this deviation is attributable to the utilization of virtualization techniques, due to 3D rendering engines. The deviation from real data is due to the use of compression methods to reduce data volumes in terms of manageability of data volume. There is a gap in understanding the quantitative impact of these deviations on ML-based perception in automated driving functions. The quantification of this impact is necessary when assessing the uncertainties of test systems~\cite{reisgys_how_2023}. 
To address this, we introduce a four-step framework quantifying the impact of image quality on ML model performance. Our findings highlight specific examples where image degradation leads to performance loss in perception-driven tasks. 

\subsection{Related Work}
The degree of realism of synthetic images is assessed using metrics like \ac{fid}, \ac{lpips} and \ac{ssim}~\cite{wang_image_2004, heusel_gans_2017, zhang_unreasonable_2018}. The metrics are used for the evaluation of generative adversarial networks (GANs), which are used to generate images as realistic as possible~\cite{valdebenito_maturana_exploration_2023, lucic_are_2018}. Another way of measuring the realism of computer-generated images is to quantify photorealism. It can be quantified by calculating a fidelity score, through metrics like Gray Level Co-occurrence Matrix (GLCM) Local Binary Patterns (LBP) based on textural and frequency properties of images. With that, a neural network is specially trained to calculate fidelity score with those metrics~\cite{duminil_comprehensive_2024, duminil_assessing_2024}. 
Both approaches can differentiate real and generated images. There are no experiments indicating the impact on \ac{ML} tasks like object detection or semantic segmentation. 
The extent to which the calculated photorealism affects the actual performance of a \ac{ML} task remains uncertain. 

The relationship between machine learning (ML) performance and image degradation due to compression has been investigated across various domains, including healthcare \cite{jo_impact_2021} and surveillance \cite{gandor_first_2022}. However, research on this topic within automotive applications remains limited. Existing studies have predominantly examined the effects of JPEG compression \cite{benbarrad_impact_2022,dejean-servieres_study_2017,bhowmik_lost_2022}, with a focus on classification and detection tasks \cite{benbarrad_impact_2022,dejean-servieres_study_2017,bhowmik_lost_2022,jo_impact_2021}. These studies have identified the SSIM as a suitable metric for assessing image quality. However, none have specifically addressed the requirements of the automotive domain or evaluated the impact of compression on tasks such as segmentation and learning-based compression methods.

\subsection{Contribution}
This work addresses key gaps by analyzing the impact of image deviations on detection and segmentation tasks using automotive datasets. The three main contributions are:
\begin{itemize}
    \item Analysis of the impact of image quality, including compression techniques and simulated environments, on object detection and segmentation tasks using automotive datasets.
    \item Evaluation of metrics to quantify image quality by their ability to  distinguish modifications.
    \item A framework for evaluating the relationship between image deviations and perception performance.
\end{itemize}
The structure of the paper is as follows: Section~\ref{sec:Problem_Statement} presents the problem statement and provides a definition of validity. Chapter~\ref{sec:Methodology} outlines the methodology, starting with the approach, followed by a description of the image quantification metrics, and introduction of metrics used to evaluate ML performance. Chapter~\ref{sec:Results} includes the experimental setup, the analysis of the considered metrics, and their correlation. Finally, the conclusion and outlook are presented in section~\ref{sec:Conclusion}.
\begin{figure*}[ht!]
    \centering
    \includegraphics[width=\linewidth]{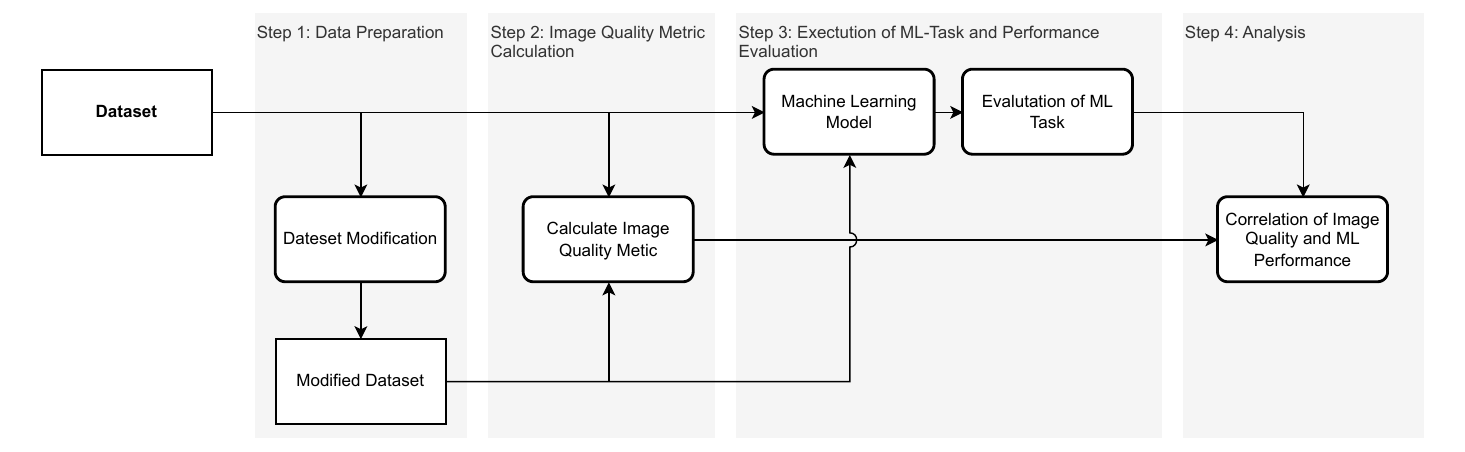}
    \caption{Concept to quantify image deviation and their influence on ML tasks according to section \ref{sec:Problem_Statement}. It consists of four steps, data preparation, image quality calculation, \ac{ML} task performance evaluation and correlation analysis.}
    \vspace{-3mm}
    \label{fig:Workflow}
\end{figure*}

\section{Problem Statement} \label{sec:Problem_Statement}

This investigation focuses on understanding the relationship between deviations in images and model performance through a structured analysis of three key aspects. First, it seeks to define how deviations between images can be recognized and quantified. Given a reference image $\mathrm{\bm{I}}_\mathrm{r}$ and a modified image $\mathrm{\bm{I}}_\mathrm{d}$, the deviation $\Delta(\mathrm{\bm{I}}_\mathrm{r},\mathrm{\bm{I}}_\mathrm{d})$ must be quantified using quality metrics $\mathcal{D}$. Second, the study explores how deviations in image quality impact the performance of a \ac{ML} model $\mathcal{M}$. For any given input image $\mathrm{\bm{I}}$, the performance is assessed using a metric $\mathrm{P}(\mathcal{M},\mathrm{\bm{I}})$. By comparing the model's performance on the reference and modified images, the change in performance $\Delta\mathrm{P}= \mathrm{P}(\mathcal{M},\mathrm{\bm{I}}_\mathrm{r})-\mathrm{P}(\mathcal{M},\mathrm{\bm{I}}_\mathrm{d})$ can be calculated. This enables the investigation of the dependency between image quality deviations $\Delta(\mathrm{\bm{I}}_\mathrm{r},\mathrm{\bm{I}}_\mathrm{d})$ and performance degradation $\Delta\mathrm{P} $, hypothesized to follow a functional relationship 
$\Delta\mathrm{P} = \bm{f}(\Delta(\mathrm{\bm{I}}_\mathrm{r},\mathrm{\bm{I}}_\mathrm{d}))$. Finally, the investigation evaluates the suitability of various metrics for quantifying image deviations and predicting performance changes. Capturing the functional relationship between  $\Delta(\mathrm{\bm{I}}_\mathrm{r},\mathrm{\bm{I}}_\mathrm{d})$ and  $\Delta\mathrm{P}$. The goal is to identify which metric $\mathcal{D}$ provides the most accurate mapping of image differences to performance impact, represented as $\bm{f}:\mathcal{D}(\mathrm{\bm{I}}_\mathrm{r},\mathrm{\bm{I}}_\mathrm{d}) \rightarrow \Delta \mathrm{P}$.

By establishing a framework for recognizing image deviations, analyzing their effects on performance, and selecting the most suitable quantification metrics, this study aims to offer a deeper understanding of how image quality influences the robustness and accuracy of \ac{ML} systems.
\subsection{Definitions}
To determine whether an image that deviates from a reference image is suitable for validation of an ML-based function, a quantification process is required. To clarify the scope of this publication, the following terms are defined:
\begin{definition}[Image Quality]\label{def:Image_Quality}
The quality of a compressed, generated, or virtualized image is defined based on the deviation between the image and a reference image captured in the real world. This deviation means the perceptual or statistical differences between the two images.
\end{definition}
According to definition \ref{def:Image_Quality} image quality metrics are all metrics, that quantify the deviation between images.
\begin{definition}[Image Modification]\label{def:Image_modification}
Image modification refers to a transformation applied to an image that introduces deviations in certain characteristics, such as compression artifacts, generative alterations, or virtualization effects, while maintaining the same semantic content.
\end{definition}
Image validity refers to the impact of loss in image quality through image modifications on the performance of \ac{ML} tasks. A reference image serves as the ground truth, against which the consistency of predictions is assessed. This ensures the reproducibility of true positives and true negatives. The preservation of false positives and false negatives is given. For instance, if a reference image contains a misclassification where a bush is erroneously detected as a car, the modified image must also produce the same misclassification to maintain image validity. Deviations in such cases could imply a loss of validity, especially if the images are used for validation. 
\begin{definition}[Validity]\label{def:validity}
Given a reference image $\bm{I}_\mathrm{r}$ and a modified image $\bm{I}_\mathrm{m}$, the modified image is considered valid for a given task if, the performance $\mathrm{P}$ of a \ac{ML} task $\mathcal{M}$ matches the output of the reference image, that is
\begin{equation}
    \mathrm{P}(\mathcal{M},\bm{I}_\mathrm{r}) == \mathrm{P}(\mathcal{M},\bm{I}_\mathrm{d}).
\end{equation}
\end{definition}

\section{Methodology}\label{sec:Methodology}
The relationship between image quality and \ac{ML} task performance is examined by systematically modifying images, assessing their quality, and analyzing the impact on model accuracy and reliability in four steps (see Figure \ref{fig:Workflow}).
In the first step, reference images are modified using three techniques: virtualization (Virtual Kitti and Virtual Kitti 2 datasets), JPEG compression (introducing compression artifacts), and Vector Quantized Generative Adversarial Networks (VQGAN) to simulate generative model distortions.
In Step 2, the image quality metrics are calculated to quantify deviations between modified and reference images. The metrics analyze the quality on pixel-level, perceptual and structural similarity, statistical feature distribution differences and differences quantified in a vector space (see Table~\ref{tab:img_metric_categories}). 
In Step 3, \ac{ML} models are evaluated on object detection and semantic segmentation tasks using metrics such as Mean Average Precision (MAP), Mean Intersection over Union (MIoU), F1 Score, and Mean Dice Coefficient (see Table~\ref{tab:ml_performance}). These metrics are applied to determine ML performance between the reference image and the modified image.  
The relationship between image quality metrics and model performance is analyzed in step 4. To identify the impact of image modifications on model accuracy and reliability, the image quality metrics are correlated with the \ac{ML} performance metrics.

\subsection{Data Preparation}
To analyze the impact of image quality on ML performance, comparable images are required. Therefore, pairs of images with identical semantic content are compared.
The image pairs are created using, three modification methods:

\textbf{JPEG Compression}: JPEG is a widely used lossy compression method for digital images. It begins by converting the image from $\mathrm{RGB}$ to $\mathrm{YCbCr}$ color space, which separates the chrominance and luminance components. The image is divided into blocks $8\times8$, and a discrete cosine transform (DCT) is applied to convert the spatial information into frequency components. These coefficients are quantized, with higher frequencies discarded, and then encoded using Huffman coding.  

\textbf{\ac{vqgan}}: VQGAN is a deep learning-based compression framework, introduced in~\cite{van_den_oord_neural_2017} and further refined in~\cite{esser_taming_2021}. It employs an encoder-decoder architecture, where the encoder transforms input data into a latent representation, which is then discretized through vector quantization by mapping the continuous latent vectors to entries in a learned codebook with a specific codebook size (CB). The decoder reconstructs the input from these quantized representations. VQGAN incorporates a perceptual loss function and an adversarial training approach to enhance the compressed representation's quality. A patch-based discriminator, trained to distinguish between real and reconstructed images, guides the generator toward producing more realistic reconstructions. This combination enables a perceptually rich codebook, improving compression efficiency and visual fidelity. 

\textbf{Virtual KITTI 1 and Virtual KITTI 2:} The Virtual KITTI datasets~\cite{gaidon_virtual_2016, cabon_virtual_2020} are characterized by its use of real video scenes from the KITTI~\cite{geiger_vision_2013} dataset that have been recreated using the Unity-Engine. The generation of these virtual scenes was achieved by reconstructing scene geometry and camera path based on real-world scenes within the Unity Engine. Virtual KITTI 1 is based on Unity 5.3/5.5. Virtual KITTI 2 employs an advanced version of Unity (2018.4 LTS) with an enhanced photorealistic rendering through lighting and shader effects in comparison to its predecessor. 

\subsection{Image Quality Metric Calculation}
\begin{table}[t]
\centering
\caption{Categorization of Image Quality Analysis Metrics}
\label{tab:img_metric_categories}
\begin{tabular}{@{}ll@{}}
\toprule
\textbf{Category}               & \textbf{Metrics}                                                               \\ \midrule
\textbf{Pixel-level} & \begin{tabular}[c]{@{}l@{}}
                                    Mean Squared Error (MSE) \\ 
                                    Peak Signal to Noise Ratio (PSNR)
                                  \end{tabular} \\ \midrule
\textbf{Perceptual}     & \begin{tabular}[c]{@{}l@{}}
                                    Learned Perceptual Image Patch Similarity (LPIPS) \\ 
                                    Structural Similarity Index (SSIM) \\ 
                                    Frechet Inception Distance (FID)
                                  \end{tabular} \\ \midrule
\textbf{Statistical} & \begin{tabular}[c]{@{}l@{}}
                                         Mutual Information (MI) \\
                                         Normalized Cross-Correlation (NCC)\\
                                         Earth Mover's Distance (EMD)\\
                                         Entropy
                                       \end{tabular} \\ \midrule
\textbf{Vector Space} & Cosine Similarity (CS)\\ \bottomrule
\end{tabular}
\vspace{-3mm}
\end{table}

Quantifying image quality is essential for evaluating the quality of reconstructions or comparisons in image processing. Metrics fall into four main categories: pixel-level fidelity, which measures direct differences between corresponding pixel values; perceptual alignment, capturing human-visual system-relevant differences; statistical similarity, which evaluates image distributions or histograms; and vector space comparisons, focusing on high-level feature representation. An overview can be seen in Tab~\ref{tab:img_metric_categories}.
\textbf{\ac{mse}}  computes the average squared difference between the pixel values, offering simplicity but failing to account for perceptual differences. \textbf{\ac{psnr}}, derived from MSE, captures pixel fidelity but is similarly limited in reflecting perceived quality.
The \textbf{\ac{ssim}}, introduced in~\cite{wang_image_2004}, improves upon these by evaluating luminance, contrast, and structure, aligning more closely with human perception. However, it is sensitive to alignment issues. \textbf{\ac{lpips}}, introduced in~\cite{zhang_unreasonable_2018}, uses deep neural network features to assess perceptual differences, achieving strong alignment with human judgment but at a computational cost.
\textbf{\ac{fid}} introduced in~\cite{heusel_gans_2017}, compares feature distributions in a latent space, providing a perceptually meaningful metric for tasks like evaluating generative models, though it demands significant resources.
\textbf{\ac{ncc}}, measures how similar two images are by comparing their intensity patterns in a way that accounts for differences in brightness and contrast.
\textbf{Mutual Information} quantifies the shared information between two images, making it valuable for image registration tasks. \textbf{Cosine Similarity} measures the angle between feature vectors, enabling effective high-level comparisons but neglecting pixel-level details and spatial structure. 
\textbf{Earth Mover's Distance (EMD)}, measures the minimal cost required to transform one probability distribution into another, which can also be applied to images.
\textbf{Entropy}, quantifies the amount of randomness or information within an image, assessing texture differences.

\subsection{ML Task and Performance Evaluation}
\begin{table}[t]
\centering
\caption{Metrics for Machine Learning Evaluation in Detection and Semantic Segmentation}
\label{tab:ml_performance}
\begin{tabular}{@{}lll@{}}
\toprule
\textbf{Task}               & \textbf{Metric}                           \\ \midrule
\textbf{Detection}          & Mean Intersection over Union (mIoU)       \\
                            & Mean Average Precision (mAP)             \\
                            & mAP Small                                \\
                            & F1 Score                                 \\ \midrule
\textbf{Semantic Segmentation} & Mean Dice Coefficient                  \\
                            & Mean Intersection over Union (mIoU)       \\
                            & Mean Pixel Accuracy                      \\ \bottomrule
\end{tabular}
\vspace{-3mm}
\end{table}

Object detection and semantic segmentation are key tasks in the automotive domain, enabling advanced perception systems for autonomous driving and driver assistance. Object detection focuses on identifying and localizing objects within a scene, while semantic segmentation provides pixel-level classification of scene components.
This contribution utilized \textbf{YOLO}~\cite{jocher_ultralytics_2023} for object detection and \textbf{Mask2Former}~\cite{cheng_masked-attention_2022} for semantic segmentation. Other models can be used in the framework if the input and output interfaces are complied with.

Evaluating \ac{ML} performance requires metrics tailored to specific tasks, as each metric captures different aspects of model behavior. A list of the metrics used in this contribution is listed in Tab.~\ref{tab:ml_performance}. For detection tasks, the \textbf{Mean Intersection over Union} (Mean IoU) measures the overlap between predicted and ground-truth bounding boxes or masks. The \textbf{Mean Average Precision} (mAP) evaluates the tradeoff between precision and recall across multiple IoU thresholds, providing a comprehensive assessment of detection performance. Specialized variants such as \textbf{mAP Small} focus on evaluating models' abilities to detect small objects. The $\mathrm{F}_1\text{-Score}$ balances precision and recall into a single value. For semantic segmentation, the \textbf{Mean Dice Coefficient}  quantifies the similarity between predicted and ground-truth segmentation masks, particularly useful for imbalanced classes. \textbf{Mean Intersection over Union} (Mean IoU) evaluates pixel-level overlap for overall segmentation accuracy. The \textbf{Mean Pixel Accuracy} measures the proportion of correctly classified pixels. It is important to note that in this paper, the ground truth is calculating the metrics is the ML performance on the reference image, as the goal is to quantify performance deviation. For example, mIoU is calculated by comparing the bounding boxes detected on the reference image with those on the modified image. 

\begin{figure*}[ht]
    \centering
    \includegraphics[width=\textwidth]{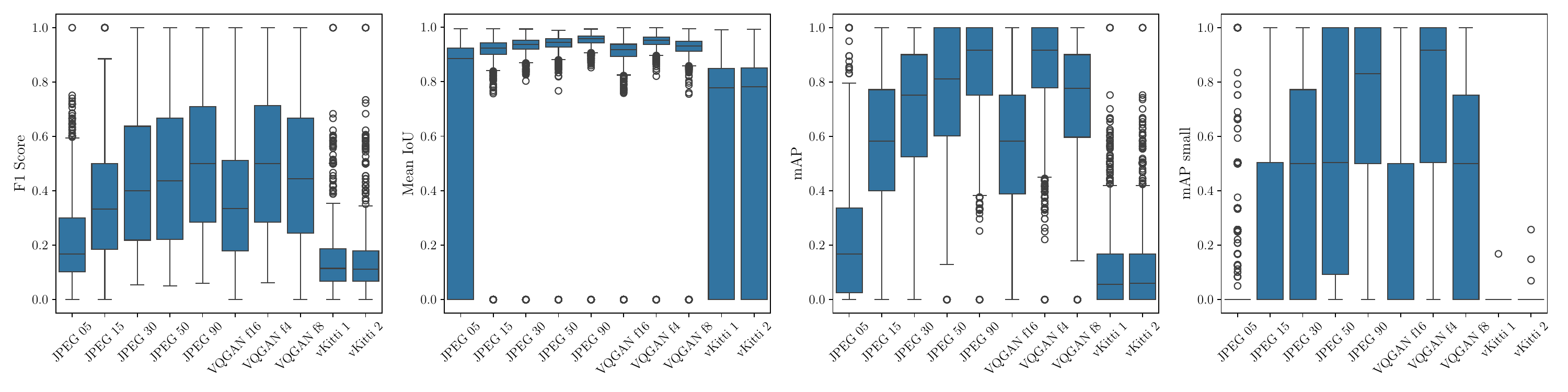} %
    \vspace{-5mm}
    \caption{Quantification of the impact of image modification on \ac{ML} performance for \textbf{object detection}. Values approaching one indicate similar performance on the modified and reference image.}
    \vspace{-3mm}
    \label{fig:box_ml_performance_plot_obj}
\end{figure*}
\begin{figure*}[ht]
    \centering
    \includegraphics[width=\textwidth]{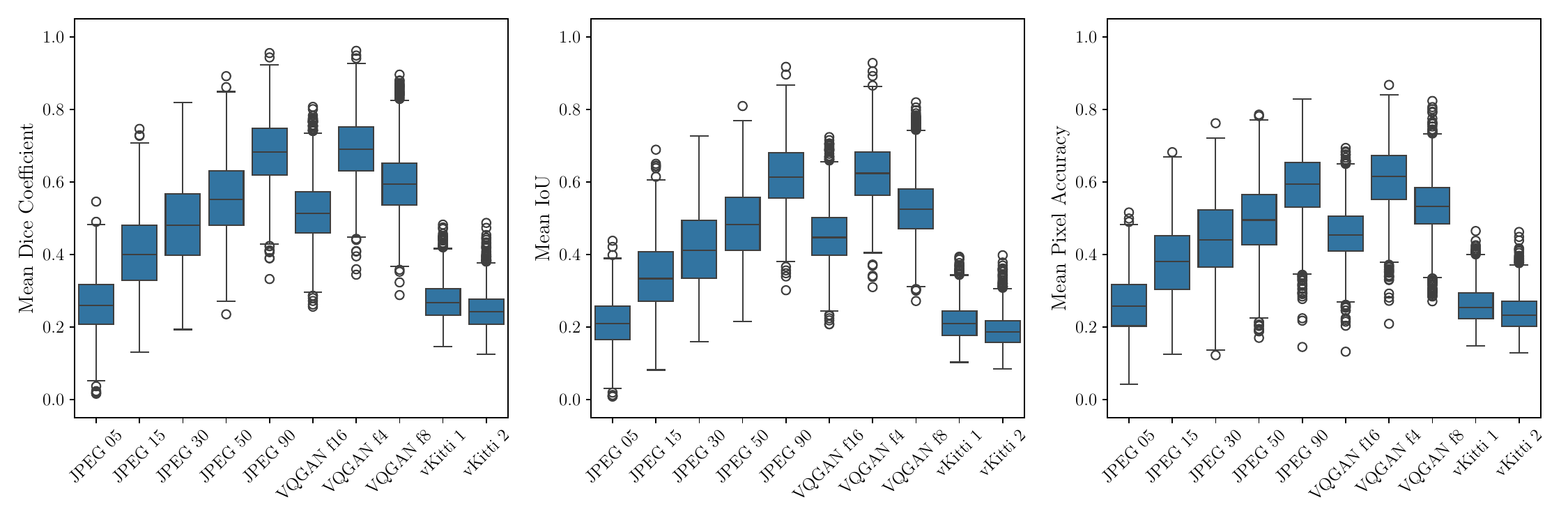} %
    \vspace{-5mm}
    \caption{Quantification of the impact of image modification on \ac{ML} performance for \textbf{semantic segmentation}.  Values approaching one indicate similar performance on the modified and reference image.}
    \vspace{-3mm}
    \label{fig:box_ml_performance_plot_sem_seg}
\end{figure*}

\begin{figure*}[ht]
    \centering
    \includegraphics[width=\textwidth]{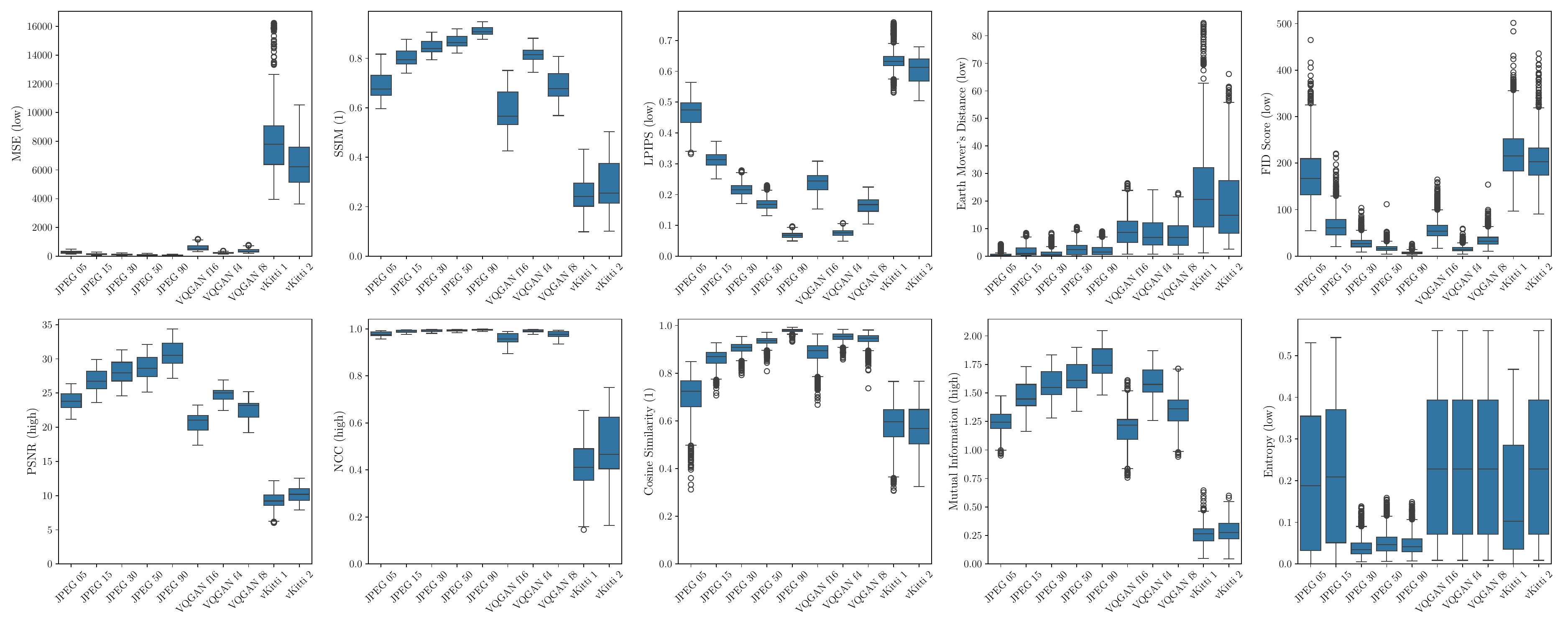} %
    \vspace{-5mm}
    \caption{Quantification of \textbf{image quality} for the modification methods. The Labels (low), (high), and (1) indicate which values are considered better.}
    \vspace{-3mm}
    \label{fig:box_quality_plot}
\end{figure*}

\subsection{Analysis}
The analysis focuses on evaluating image quality metrics in terms of their ability to differentiate between image modifications. Additionally, the relationship between image quality and machine learning performance is examined to understand the potential influence of image quality on model effectiveness. A correlation analysis is conducted to quantify the relationship between image quality metrics and machine learning performance metrics.
\section{Results}\label{sec:Results}
\begin{table}[t]
\caption{Overview of compression factors JPEG and \ac{vqgan} on the KITTI dataset.}
\label{Tab:comp_factor}
\centering
\begin{tabular}{l c c }
\toprule
 Method & Setting & Compression Factor \\ \midrule
JPEG  & $\mathrm{Quality}=90$ &$6.03\pm0.43$ \\ 
JPEG  &$\mathrm{Quality}=50$ &$15.81\pm1.28$ \\ 
JPEG  &$\mathrm{Quality}=30$ &$21.55\pm1.88$ \\ 
JPEG  & $\mathrm{Quality}=15$ &$32.835\pm2.94$ \\ 
JPEG  & $\mathrm{Quality}=5$ &$59.56\pm3.51$ \\ 
VQGAN  & f4, $\mathrm{CB=2^{13}}$ &$29.53\pm0$ \\ 
VQGAN & f8, $\mathrm{CB=2^{14}}$ &$109.71\pm0$ \\ 
VQGAN  & f16, $\mathrm{CB=2^{13}}$ &$438.85\pm0$  \\ \bottomrule
\end{tabular}
\vspace{-3mm}
\end{table}

\subsection{Experimental Setup}\label{sec:Dataset}
The reference dataset consists of 2126 images, based on 5 sequences of the Kitti dataset, which are also available as 3D simulated images within the virtual Kitti 1 and 2 datasets.   
For JPEG compression, the \textit{python-pillow} package is used. For VQGAN the python implementation and pre-trained models of~\cite{rombach_high-resolution_2021} are used. An overview of both compression methods regarding their compression factors can be seen in Tab.~\ref{Tab:comp_factor}. 
For semantic segmentation, we used theMask2Former with a pre-trained model  (\textit{mask2former-swin-small-cityscapes-panoptic}) from Facebook~\cite{cheng_masked-attention_2022}. For the object detection, a pre-trained model for YOLO11m from Ultralytics is used~\cite{jocher_ultralytics_2023}. 

\subsection{Evaluation of Image Quality and ML Performance Metrics}
Figures \ref{fig:box_ml_performance_plot_obj} and \ref{fig:box_ml_performance_plot_sem_seg} show the distribution of ML performance metrics for image modifications. Values approaching one indicate minimal performance deviation. Across all evaluated ML performance metrics, values remain consistently below one. This indicates a general degradation in performance due to image modifications. For the object detection (see Figure~\ref{fig:box_ml_performance_plot_obj}), the $\mathrm{F}_1\text{-Score}$ exhibits the highest deviation. A decline in the $\mathrm{F}_1\text{-Score}$ leads to an increased number of missed detections or misclassifications. Small object detection accuracy is most affected, as shown by the high variance in mAP small. The low values in Figure~\ref{fig:box_ml_performance_plot_sem_seg} indicate a degradation in segmentation performance. A reduced Mean Dice Coefficient and Mean IoU, suggest a small overlap between predicted and ground truth segmentation masks. The low Mean Pixel Accuracy indicates misclassifications. Our results show that modifications to input data, such as compression or simulation within 3D environments, directly impact \ac{ML} performance. Even minimal changes $\Delta(\mathrm{\bm{I}}_\mathrm{r},\mathrm{\bm{I}}_\mathrm{d})$ can lead to performance degradation $\Delta\mathrm{P}$. Therefore, all modifications investigated in this paper lead to invalid images according to definition \ref{def:validity}. Introducing confidence intervals can provide a more nuanced image validity definition for testing automated driving functions.

Figure \ref{fig:box_quality_plot} depicts the image quality metrics computed for each type of modification. The distribution of the metrics is visualized using box plots. Bigger boxes indicate higher variations. The metrics are assessed based on their ability to separate modifications, variance behavior, and sensitivity to compression. SSIM, LPIPS, and Mutual Information effectively separate modifications, forming distinct clusters within one modification method indicated by small variations. PSNR, FID, Cosine Similarity, and Mutual Information show cluster overlap in combination with a higher variation compared to SSIM, LPIPS, and Mutual Information. This indicates a slightly lower performance in separating the image modifications. MSE and NCC have no sensitivity to compression and show high variations in assessing the quantification of the difference between simulated and real images. Earth Mover's Distance can distinguish between the modification methods. However, within the modification method, the clusters are overlapping. Entropy has the highest variance across all metrics. It cannot separate between the modification methods. 
Overall, LPIPS provides the highest separability with minimal overlap, due to its low variation in value clusters across all modifications. 

\subsection{Correlation Analysis}
\begin{table*}[ht!]
\centering
\caption{Correlation of Object Detection and Image Quality Metrics}
\label{tab:correlation_matrix_Object_detection}
\resizebox{\textwidth}{!}{%
\begin{tabular}{lcccccccccc}
\toprule
\makebox[3cm][l]{\textbf{Metric}} & \textbf{MSE} & \textbf{PSNR} & \textbf{SSIM} & \textbf{NCC} & \textbf{LPIPS} & \textbf{Cosine Similarity} & \textbf{\begin{tabular}[c]{@{}c@{}}Earth Mover's\\ Distance\end{tabular}} & \textbf{Mutual Information} & \textbf{FID Score} & \textbf{Entropy} \\ \midrule
\textbf{F1 Score} & -0.35 & 0.37 & 0.39 & 0.36 & \textbf{-0.48} & 0.45 & -0.20  & 0.41 & -0.44 & -0.08 \\
\textbf{Mean IoU} & -0.58 & 0.61 & 0.63 & 0.59 & \textbf{-0.65} & 0.62 & -0.36  & 0.65 & -0.60 & -0.13\\
\textbf{mAP} & -0.54 & 0.58 & 0.62 & 0.55 & \textbf{-0.74} & 0.69 & -0.31  & 0.65 & -0.69 & -0.14 \\
\textbf{mAP small} & -0.34 & 0.43 & 0.46 & 0.37 & \textbf{-0.55} & 0.49 & -0.21  & 0.47 & -0.49 & -0.12\\
\bottomrule
\end{tabular}%
}
\end{table*}
\begin{table*}[ht!]
\centering
\caption{Correlation of Semantic Segmentation and Image Quality Metrics }
\label{tab:correlation_matrix_semantic_seg}
\resizebox{\textwidth}{!}{%
\begin{tabular}{lcccccccccc}
\toprule
\makebox[3cm][l]{\textbf{Metric}}& \textbf{MSE} & \textbf{PSNR} & \textbf{SSIM} & \textbf{NCC} & \textbf{LPIPS} & \textbf{Cosine Similarity} & \textbf{\begin{tabular}[c]{@{}c@{}}Earth Mover's\\ Distance\end{tabular}} & \textbf{Mutual Information} & \textbf{FID score} & \textbf{Entropy} \\ \midrule
\textbf{Mean Dice Coefficient} & -0.55 & 0.58 & 0.61 & 0.57 & \textbf{-0.83} & 0.76 & -0.28 & 0.66 & -0.75 & -0.12\\
\textbf{Mean IoU} & -0.56 & 0.59 & 0.62 & 0.57 & \textbf{-0.87} & 0.77 & -0.28 & 0.67 & -0.75 & -0.12\\
\textbf{Mean Pixel Accuracy} & -0.54 & 0.57 & 0.60 & 0.56 & \textbf{-0.80} & 0.74 & -0.28 & 0.65 & -0.72 & -0.12\\
\bottomrule
\end{tabular}%
}
\vspace{-3mm}
\end{table*}
Tables~\ref{tab:correlation_matrix_Object_detection} and~\ref{tab:correlation_matrix_semantic_seg} show the correlation between image quality metrics and \ac{ML} model performance for object detection and semantic segmentation. Positive correlations indicate alignment between higher quality metric values and better model performance. Negative correlations reflect an inverse relationship. Correlation values near zero suggest weak or no significant relationships.
The analysis highlights two key findings: First, semantic segmentation tasks exhibit stronger correlations with image quality metrics compared to object detection tasks. Second, the \ac{lpips} metric consistently demonstrates the highest correlation across both tasks, making it an indicator of performance changes.
For object detection (see Table~\ref{tab:correlation_matrix_Object_detection}), image quality metrics generally correlate more strongly with Mean IoU, MAP, and MAP small, while associations with the $\mathrm{F}_1\text{-Score}$ are more moderate. Among these, \ac{lpips} stands out with the strongest correlation, whereas Entropy shows the weakest correlation.
Similarly, in semantic segmentation (see Table~\ref{tab:correlation_matrix_semantic_seg}), \ac{lpips} demonstrates the strongest correlation with all performance metrics, whereas Entropy remains the least predictive. Semantic segmentation shows stronger correlations with image quality metrics than object detection. This is evident from the higher LPIPS correlation for semantic segmentation (0.87) versus object detection (0.74). 

\subsection{Discussion}
Among the metrics evaluated, the \ac{lpips} metric emerged as the most effective. It exhibited the highest sensitivity to modifications and maintained low variance, ensuring reliable differentiation between methods of modification $\Delta(\mathrm{\bm{I}}_\mathrm{r},\mathrm{\bm{I}}_\mathrm{d})$. Additionally, it showed the strongest correlation with performance degradation, affirming its ability to predict the impact of input deviations on ML tasks $\Delta\mathrm{P} = \bm{f}(\Delta(\mathrm{\bm{I}}_\mathrm{r},\mathrm{\bm{I}}_\mathrm{d}))$. The metric provided robust separation across the tested modification methods, highlighting its capability to distinguish even subtle changes.

\section{Conclusion and Outlook}\label{sec:Conclusion}
This study investigated the impact of input data modifications, such as compression, generative model artifacts, and simulation-induced distortions, on machine learning performance of image-based perception for automated driving. Our findings demonstrate that, even minor alterations to input images can lead to significant performance degradation. Among the evaluated quality metrics, LPIPS proved to be the most effective predictor of performance degradation, exhibiting high sensitivity, low variance, and strong correlation with observed performance drops across various modification methods. This underscores LPIPS's capability to quantify the impact of input deviations on ML tasks.
In practice, our framework is suitable for quantifying the uncertainty of test systems or test data. It effectively identifies which metric best correlates with the ML performance of a System under Test (SuT). In our experiments, LPIPS demonstrated the strongest correlation, making it a valuable metric for predicting ML performance.

Moving forward, future research will focus on establishing  confidence intervals or weighted deviations based on relevance to the automated driving system. The findings of how the quantification of image quality can be integrated into a validation process for simulation environments need to be investigated.
Additionally, the investigation can be expanded to simulation environments with varying levels of photorealism. 
\bibliography{references.bib}
\end{document}